\documentclass[letterpaper]{article}
\usepackage{imav}
\usepackage{flushend}
\usepackage{amsmath}
\usepackage{amssymb}
\usepackage{times}
\usepackage{graphicx}
\usepackage{layout}
\usepackage{wasysym}
\usepackage{xcolor}
\usepackage{subcaption}
\usepackage{comment} 

\title{LiDARFlow: Real-Time Panel-Based MAV Guidance in Unknown Environments}
\author{João Machado$^{1}$\thanks{Email address: joao.machado@enac.fr, zeynep.bilgin@irit.fr , matthieu.verdoucq@enac.fr, murat.bronz@enac.fr}, Zeynep Bilgin$^{2}$, Matthieu Verdoucq$^{1}$, Murat Bronz$^{1}$ \\
$^1$ENAC, École Nationale de l’Aviation Civile, Université de Toulouse, Toulouse, France\\
$^2$CNRS, Centre national de la recherche scientifique, France}

\begin{document}
\maketitle 
\thispagestyle{empty} 

\begin{abstract}
This paper presents a guidance algorithm for micro aerial vehicles operating in unknown, cluttered environments using only onboard sensing. The method is based on a panel formulation originally derived from aerodynamic potential-flow theory and generates smooth, collision-free guidance vectors from locally perceived obstacles. The approach is extended to unknown environments by constructing and updating the obstacle representation online from onboard LiDAR measurements. The resulting obstacle-avoidance field is integrated with a nominal guiding vector field to produce the final control input. The system is experimentally validated in indoor flight tests under two scenarios: waypoint navigation and directional guidance. In both cases, the vehicle successfully completes its task while avoiding all obstacles in real time using only onboard perception. The results demonstrate that the method is computationally lightweight and suitable for onboard implementation, with point-cloud processing identified as the main practical limitation. These results support the feasibility of lightweight onboard guidance in unknown environments.


\end{abstract}


\section{Introduction} \label{section:introduction}

Advances in technology have made autonomous micro aerial vehicles (MAVs increasingly important across a wide range of applications such as search and rescue and disaster relief. Thanks to their small size and low weight, MAVs can reach locations inaccessible to humans, be deployed in large numbers, and provide vantage points that would otherwise be unattainable. In many of these scenarios, MAVs must operate in unknown environments for which no prior map is available. For example, in forests or disaster zones, pre-mapping all obstacles is often infeasible. Consequently, onboard sense-and-avoid capability is essential for mission success. Operating reliably under these conditions requires a robust guidance system that is lightweight enough to run onboard the MAV.

\begin{figure}[!t]
    \centering
    \includegraphics[width=\columnwidth]{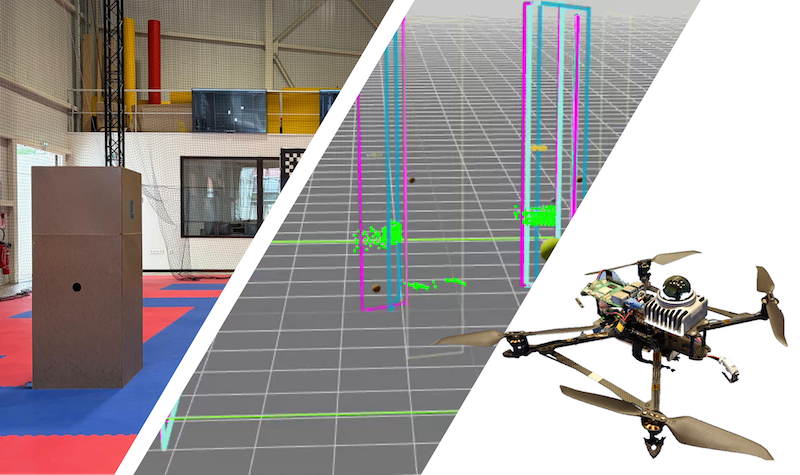}
    \caption{Stage overview of the project. The obstacles, represented on the left, are reconstructed with LiDAR data, as illustrated in the middle. The model of the drone used during the experiments can be seen on the write. It runs onboard with a Raspberry Pi running the nominal guidance and \textit{PGFlow} algorithms, producing velocity commands that are passed to the onboard paparazzi micro-controller.}
    \label{fig:catchy_fig}
\end{figure}

In many of these applications, MAVs are deployed by personnel who may not have extensive experience in piloting aerial robots. This motivates the development of guidance systems that reduce operator workload by abstracting low-level control tasks such as continuous obstacle avoidance, allowing users to focus on high-level mission objectives. Such capabilities are particularly valuable in time-critical missions where trained operators may be limited or unavailable.

\subsection{Related work}
Existing approaches to UAV guidance in unknown environments fall broadly into
two categories: optimization-based methods and reactive methods. The first
formulates guidance as a trajectory optimization problem and has converged on
two competing paradigms. Hard-constraint corridor and mixed-integer (MIQP)
planners such as FASTER~\cite{faster2019}, MADER~\cite{mader2022},
HDSM~\cite{hdsm2024}, DYNUS~\cite{dynus2025}, and SANDO~\cite{sando2026} guarantee
collision-free trajectories. Whereas soft-constraint gradient-based
planners like EGO-Planner~\cite{egoplanner2021}, EGO-Swarm~\cite{egoswarm2021}, and
SUPER~\cite{super2025} embed collision risk as a penalty term to achieve
faster computation at the cost of any formal safety guarantee. These methods
typically pair a global planner with a local planner that compute and refine a
safe path online. Their main drawbacks are that they generally rely on
constructing and maintaining a map of the environment and that solving the
underlying optimization is computationally demanding, which can limit the achievable replanning rate on the constrained hardware of a small MAVs.\\

The second category comprises reactive methods, which compute control commands
directly from instantaneous sensor data. Recent reactive guidance for UAVs in
cluttered, unknown environments has developed along three main lines. The first
is control barrier function (CBF) safety filters operating directly on LiDAR or
point-cloud data, exemplified by the composite CBF filter of Harms et
al.~\cite{compositecbf2025}, which steers an onboard LiDAR quadrotor through
unknown indoor and outdoor clutter while handling thousands of constraints. The
second is velocity obstacle and ORCA schemes, increasingly used together with
model-predictive control~\cite{geoprovo2024} and reinforcement
learning~\cite{rvodrl2025} for multi-agent and dynamic obstacle avoidance. The
third is potential field, guidance vector field methods, which
remain heavily used but are increasingly embedded as components within
optimisation frameworks as local planners rather than deployed as standalone controllers. 
Because they act on live sensor measurements rather than on a global map, reactive
methods offer low computational cost, short reaction latency, and map free
operation, making them well suited to the limited onboard resources of an MAV. 
Their classical weaknesses lie in completeness, optimality, oscillation, and the
absence of continuous time safety guarantees. Potential field methods in
particular remain widely used but are well known to suffer from local minima,
where competing attractive and repulsive forces can trap the vehicle short of
its goal. In \cite{apfmpc2025} this is mitigated by embedding the field within
an MPC formulation~. 

The guidance method proposed in this work superposes the velocity vector fields generated by two algorithms. The nominal motion is generated using the convergent guiding vector field proposed in \cite{Machado2026}, while obstacle avoidance is achieved using the harmonic potential field approach of \cite{bilgin_urban_2024,fahimi_obstacle_2009}. Both algorithms are individually free of local minima under their respective assumptions, while being computationally light. Although no formal proof exists that their superposition preserves this property, \cite{Machado2026} provides an intuitive argument for why undesired local minima are unlikely to arise. The practical effectiveness of the proposed guidance strategy is further demonstrated through the experimental results presented in section~\ref{section:experimental}.

The harmonic potential field approach of \cite{bilgin_urban_2024} assumes complete \textit{a priori} knowledge of obstacle locations and geometries. Although several studies have extended potential flow guidance to unknown environments, important limitations remain. In \cite{cruz2012}, obstacle avoidance was demonstrated for an aerial vehicle operating in an unknown environment, but only in hardware-in-the-loop simulation. The related panel method framework proposed in \cite{velagic2020} was validated in unknown environments using a ground robot rather than an aerial platform. More recently, \cite{rudzki2024} considered potential flow guidance for UAVs in unknown environments, but validation was limited to simulation. This work addresses these limitations by proposing an onboard lightweight system that integrates LiDAR-based obstacle detection with a collision-free guidance strategy that operates without prior knowledge of the environment and is validated through real flight experiments.

\subsection{Contributions}

This paper presents a computationally lightweight onboard guidance system for MAV navigation in unknown environments using only onboard LiDAR sensing. The proposed approach combines a convergent guiding vector field with harmonic potential-flow obstacle avoidance and operates without prior knowledge of obstacle locations or geometries. The system is validated experimentally in two representative scenarios: waypoint navigation and commanded-direction flight.

The main contributions are:
\begin{itemize}
    \item A guidance system that autonomously avoids obstacles, reducing operator workload by removing the need for manual obstacle avoidance;
    \item Full onboard integration of a computationally light functional LiDAR obstacle detection system and collision-free guidance algorithm.
\end{itemize}

An overview of the project can be seen in Fig. \ref{fig:catchy_fig}.

\section{Problem formulation}
The objective is to integrate LiDAR dynamic obstacle detection with a collision-free guidance algorithm. To validate the proposed system, two representative flight scenarios are considered in an unknown environment with unknown and dynamic obstacles. In both cases, the MAV has no prior map of the environment and must continuously detect and avoid obstacles in real time using onboard sensing.

The first scenario considers waypoint navigation. The MAV is initially located at an arbitrary position and is commanded by a human operator to reach a goal waypoint. Upon reaching this location, the operator may update the mission by assigning a new waypoint (e.g., returning to the starting location). The vehicle must therefore repeatedly track user-defined waypoints while autonomously handling all obstacle avoidance.

The second scenario considers directional guidance. Instead of fixed waypoints, the operator specifies a desired direction of motion that can be updated during flight. The MAV is required to continuously progress along the commanded direction while autonomously avoiding obstacles encountered along its path.

Together, these two scenarios evaluate the system under operational modes where operators do not need extensive piloting expertise, relying on autonomous obstacle avoidance to safely execute missions in cluttered and unknown environments.

\section{System architecture}\label{section:method}
The guidance algorithm presented in \cite{Machado2026} is used to generate the drone's nominal motion. In addition to its low computational cost, the algorithm is particularly well suited to the proposed approach because it is based on the superposition of vector fields. The original formulation does not address obstacle avoidance and so it is augmented with an obstacle-avoidance module described in the next section, which likewise relies on the superposition of velocity vector fields.

\begin{figure*}[t]
    \centering
    \includegraphics[width=1.0\textwidth]{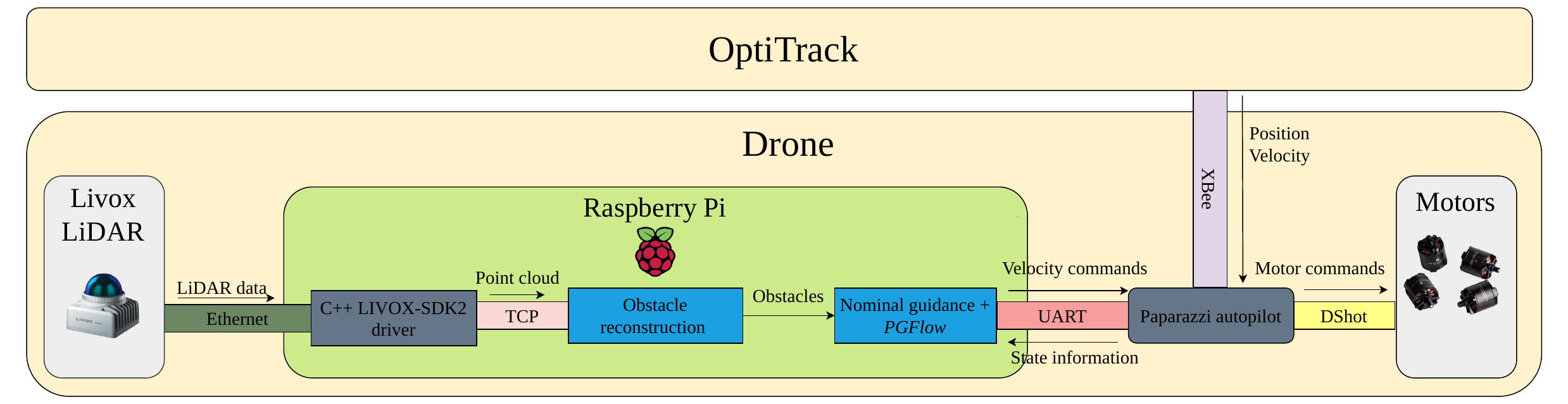}
    \caption{System architecture and respective data flow. LiDAR sends captured data to the Raspberry Pi where it is then processed by the different modules. The guidance algorithm generates the velocity commands sent to the autopilot. The autopilot fuses the vehicle's position and velocity captured by OptiTrack into the full vehicle's state, sending it to the Raspberry Pi. It also convert the velocity commands into the appropriate motor commands.}
    \label{fig:architecture}
\end{figure*}

Obstacles are detected using an onboard Livox MID-360 LiDAR sensor, which produces dense three-dimensional point clouds which are processed into usable obstacles and sent to the guidance algorithm. The full system architecture can be observed in Fig. \ref{fig:architecture}.

\subsection{Nominal motion}

To apply the algorithm in \cite{Machado2026}, it is necessary to compute the normal and tangent vectors to the point and line manifolds. The position of a point can be expressed in a three-dimensional Cartesian space by:

\begin{equation*}
\begin{aligned}
    x - p_x = 0 && y -p_y = 0 && z - p_z = 0
\end{aligned}
\end{equation*}

\noindent where $(p_x, p_y, p_z) \in \mathbb{R}^3$. Defining $F_p(x,y,z) := \begin{bmatrix}
    F_{p,1} & F_{p,2} & F_{p,3}
\end{bmatrix}^\intercal =\begin{bmatrix}
    x-p_x & y-p_y & z-p_z
\end{bmatrix}^\intercal$, the normal and tangent manifold vectors can be obtained \cite{Machado2026}:

\begin{equation*}
    \begin{aligned}
        \nabla F_{p,1} := \begin{bmatrix}
            1 & 0 & 0
        \end{bmatrix}^\intercal && \nabla F_{p,2} :=\begin{bmatrix}
            0 & 1 & 0        
    \end{bmatrix} ^ \intercal \\
    \nabla F_{p,3} : =\begin{bmatrix}
        0 & 0 & 1
    \end{bmatrix} ^\intercal && \tau_p := \begin{bmatrix}
        0 & 0 & 0
    \end{bmatrix} ^\intercal
    \end{aligned}
\end{equation*}

Accordingly, a line parallel to $xOy$ can be defined as:

\begin{equation*}
    \begin{aligned}
        &z - z_0 = 0\\
        &-\sin{\alpha}(x -x_0) + \cos{\alpha}(y-y_0) = 0
    \end{aligned}
\end{equation*}

\noindent where $(x_0, y_0, z_0) \in \mathbb{R}^3$ is an offset given to the line and $\alpha$ the desired angle between the line and the $x$-axis. Defining $F_l(x,y,z) := \begin{bmatrix}
    F_{l,1} & F_{l,2}
\end{bmatrix}^\intercal = \begin{bmatrix}
    z-z_0 & -\sin{\alpha}(x -x_0) + \cos{\alpha}(y-y_0)
\end{bmatrix}^\intercal$, the normal and tangent manifold vectors can be obtained:

\begin{equation*}
\begin{gathered}
\begin{aligned}
\nabla F_{l,1} &:=
\begin{bmatrix}
0 & 0 & 1
\end{bmatrix}^{\intercal}
\qquad
\nabla F_{l,2} &:=
\begin{bmatrix}
-\sin\alpha & \cos\alpha & 0
\end{bmatrix}^{\intercal}
\end{aligned}
\\[1ex]
\tau_l :=
\begin{bmatrix}
\cos\alpha & \sin\alpha & 0
\end{bmatrix}^{\intercal}
\end{gathered}
\end{equation*}

These can be directly introduced into the guidance algorithm to compute a velocity command prior to obstacle detection, $V$:

\begin{equation}
\label{eq:V}
\begin{aligned}
    &\text{Point:} \quad V = -v_d\frac{k_c}{3}\sum_{i=1}^3F_{p,i}\frac{\nabla F_{p,i}}{\|\nabla F_{p,i}\|} \\
    &\text{Line: }\quad V = v_d\left(\frac{\tau_l}{\|\tau_l\|}-\frac{k_c}{2}\sum_{i=1}^2F_{l,i}\frac{\nabla F_{l,i}}{\|\nabla F_{l,i}\|}\right)
\end{aligned}
\end{equation}

\noindent where $v_d$ is the desired velocity along the tangent vector\footnote{for the point case, since the tangent vector is \textbf{0}, this can be simply treated as a gain and the total convergence gain becomes $k'_c := v_dk_c$.} and $k_c$ is an arbitrary convergence gain. Note that for the line case, the drone is not converging to a particular line path. It is following the direction of the line, and so \eqref{eq:V} reduces to:

\begin{equation*}
  \text{Line:} \quad  V = v_d \frac{\tau_l}{\|\tau_l\|}
\end{equation*}

See \cite{Machado2026} for a detailed derivation of the algorithm.

\subsection{Obstacle avoidance with PGFlow}

\begin{figure}[t]
    \centering
    \includegraphics[width=1.0\linewidth, page=1]{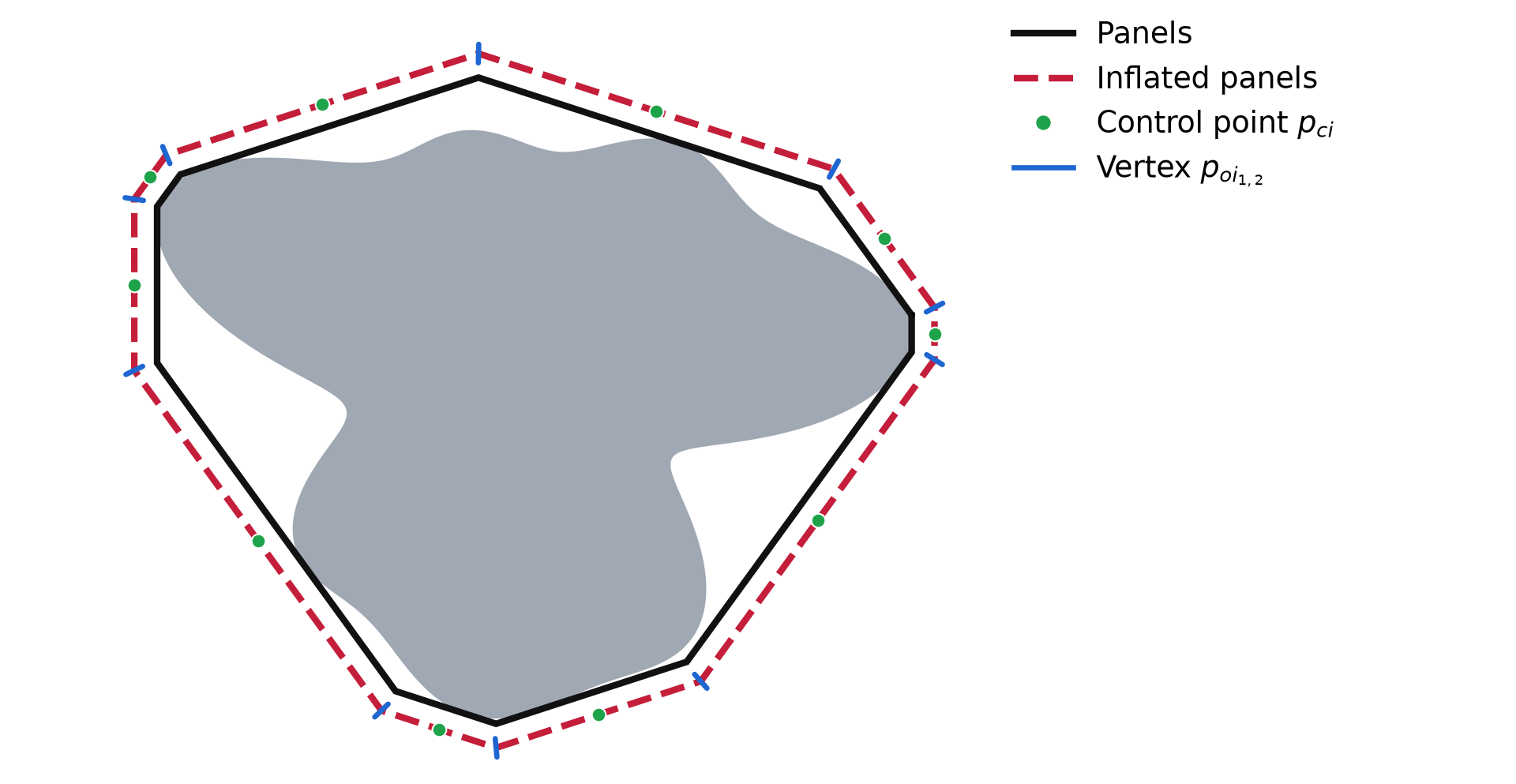}
    \caption{The obstacle (gray) is decomposed into panels (black) that act as a line source for the guidance of the drone. The panels are inflated  (red) by at least half the drone's size so that collision avoidance is guaranteed. The control points  (green) are located at each panel's half length.}
    \label{fig:obstacle}
\end{figure}

For obstacle avoidance, the PGFlow algorithm \cite{fahimi_obstacle_2009,pgflow_jgcd2024} is employed. PGFlow is a fluid dynamics inspired method that models obstacles as collections of one dimensional panels in a two-dimensional plane. Each panel is represented by a line source (or vortex, depending on the formulation), generating an irrotational velocity field around the obstacle. Fig. \ref{fig:obstacle} illustrates how an obstacle is decomposed into panels.

Since obstacle avoidance is generally more critical in the horizontal plane, only the $x$ and $y$ components of the repulsive field are considered. Although the method can be extended to three dimensions, this significantly increases computational complexity. Vertical collision avoidance can instead be handled through a simpler repulsion mechanism acting along the $z$-axis. 

Given a panel $i$ with vertices $p_{oi_1}\rightarrow(x_{oi_1}, y_{oi_1})$ and $p_{oi_2} \rightarrow (x_{oi_2}, y_{oi_2})$, its control point is defined as the centre point of the panel.

$$
p_{ci} := \frac{(p_{oi_2}-p_{oi_1})}{2} +p_{oi_1} \rightarrow (x_{ci}, y_{c_i})
$$

The velocity command given to the drone is:

\begin{equation*}
\begin{aligned}
    \dot{x} = V_x - \sum_{i=1}^m\frac{\lambda_i}{2\pi}\int_i \frac{\partial}{\partial x}\ln{R_i}\mathrm{d}l_i\\
    \dot{y} = V_y - \sum_{i=1}^m\frac{\lambda_i}{2\pi}\int_i \frac{\partial}{\partial y}\ln{R_i}\mathrm{d}l_i
\end{aligned}
\end{equation*}

where $V=\begin{bmatrix}V_x & V_y\end{bmatrix}^{\intercal}$ is the velocity command prior to \textit{PGFlow}, $m$ is the total number of panels, $\lambda_i$ is the strength of panel $i$, $R_i$ is the distance from the drone to a point on panel $i$, and $\mathrm{d}l_i$ is the differential length element along panel $i$.

The panel strengths are computed such that the normal velocity at each panel control point satisfies $V_i \geq 0$, preventing trajectories from penetrating the obstacle boundary and thereby producing a repulsive effect:

\begin{equation*}
\begin{aligned}
    &\hspace{-10mm}\text{For panel $i$ :}\\
    &\frac{\lambda_i}{2} + \sum_{i \neq j}^m\frac{\lambda_j}{2\pi}I_{ij} = -V_i + V(x_{ci}, y_{ci}) \cdot \overrightarrow{n_i}
\end{aligned}
\end{equation*}

\noindent where $I_{ij}$ is the panel interaction matrix \cite{fahimi_obstacle_2009}, and $\mathbf{n}_i$ is the outward-pointing normal vector of panel $i$. See \cite{fahimi_obstacle_2009} and \cite{bilgin_urban_2024} for a detailed derivation and discussion of the algorithm.

To account for the physical dimensions of the drone, obstacles must be inflated by at least half the characteristic size of the vehicle. Under these conditions, the method guarantees that the drone does not collide with the obstacles. This is also validated empirically on section \ref{section:experimental}.

In the proposed implementation, the obstacle panels used by PGFlow are not known a priori. They are extracted from onboard LiDAR data.

\subsection{LiDAR-based obstacle extraction}

Obstacle detection is performed using a Livox MID-360 solid-state LiDAR operating at approximately 10 Hz. At each scan, the sensor produces a high-density point cloud in the LiDAR local coordinate frame. The objective of the perception layer is to extract a compact set of obstacles that can be directly used by PGFlow.

The LiDAR scans are acquired onboard by a Raspberry Pi using a C++ driver based on Livox-SDK2 and published through a ZeroMQ stream. The drone's position is provided by the OptiTrack via XBee radio modem fused onboard with the drone's IMU. The LiDAR scans are then associated with the corresponding state estimate and transformed into a common ENU frame.

To reduce computational cost, the point cloud is first downsampled using a voxel grid with voxel side length $s = 0.2$ meters. Each point $p_i=(x_i,y_i,z_i)$ is assigned to a voxel index

\begin{equation*}
K_i = 1000003\,k_{x_i} + 1001\,k_{y_i} + k_{z_i},
\end{equation*}

where

\begin{equation*}
k_{x_i}=\left\lfloor\frac{x_i}{s}\right\rfloor,\qquad
k_{y_i}=\left\lfloor\frac{y_i}{s}\right\rfloor,\qquad
k_{z_i}=\left\lfloor\frac{z_i}{s}\right\rfloor.
\end{equation*}

All points assigned to the same voxel are replaced by their centroid. The resulting cloud is then cropped to the region of interest, which corresponds to the size of the used portion of the arena, 

\begin{equation*}
x,y \in [-3,3]~\mathrm{m}, \qquad z \in [0.5,3]~\mathrm{m},
\end{equation*}

\noindent and points closer than $0.2~\mathrm{m}$ to the sensor are discarded to remove returns from the vehicle body or from very near-field artifacts.

The filtered points are projected onto the horizontal plane and grouped into clusters. Candidate obstacle faces are extracted from each cluster using iterative RANSAC line fitting, allowing multiple planar faces to be identified within a single cluster. The extracted line segments are filtered according to geometric consistency, including minimum length, point support, continuity, and thickness, to reject spurious detections and over-extended fits. 

The remaining segments are associated with a temporal segment map, where compatible observations update existing segments and unmatched observations initialize new ones. Repeated detections increase the confidence of a segment, while temporary occlusions are handled by maintaining segments for a limited duration before removal. 

\begin{figure}
  \begin{subfigure}[b]{0.23\textwidth}
    \includegraphics[width=\linewidth]{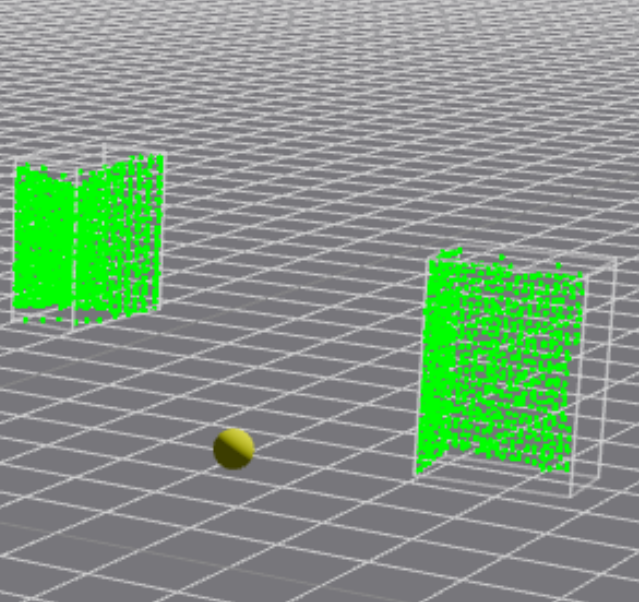}
    \caption{}
    \label{fig:pointcloud}
  \end{subfigure}
  \begin{subfigure}[b]{0.23\textwidth}
    \includegraphics[width=\linewidth,page=2]{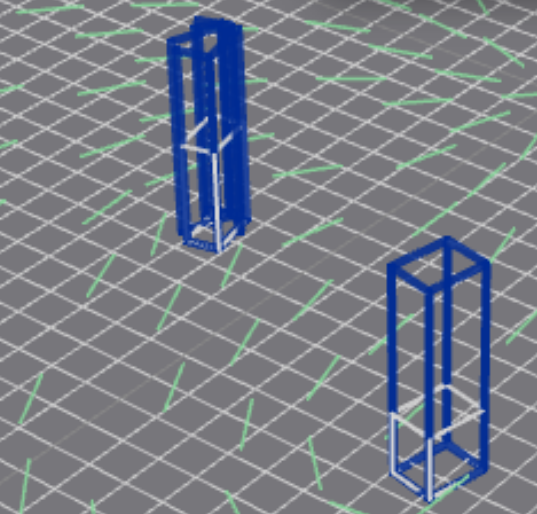}
    \caption{}
    \label{fig:obstacles}
  \end{subfigure}
  \caption{Raw pointcloud generated by the LiDAR (a), in green, and resulting obstacle segments, in blue, after processing by RANSAC (b).}
  \label{fig:lidar}
\end{figure}

\begin{figure*}[t]
    \centering
    \includegraphics[width=1.0\textwidth]{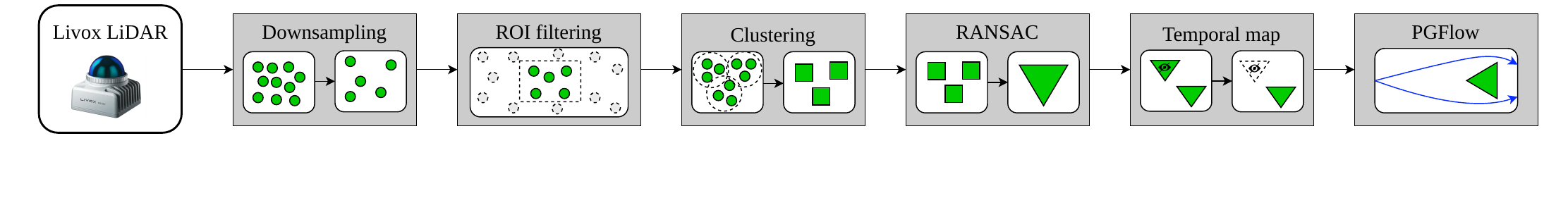}
    \caption{LiDAR's obstacle detection and processing pipeline. The raw data is captured by the Livox LiDAR sensor and then processed along the pipeline until it can be used by \textit{PGFlow} to generate the appropriate velocity commands for the drone.}
    \label{fig:LiDARflow}
\end{figure*}

The perception layer outputs an obstacle-indexed collection of segments, each defined by its endpoints, tangent direction, outward normal, and confidence. This compact representation is passed directly to the PGFlow guidance algorithm for real-time obstacle avoidance. The raw LiDAR generated pointcloud can be observed in Fig. \ref{fig:pointcloud}. Fig. \ref{fig:obstacles} shows the obstacle segments generated from that same pointcloud. The full LiDAR obstacle detection and processing pipeline can be observed in Fig. \ref{fig:LiDARflow}.

\section{Results} \label{section:experimental}

\subsection{Experimental setup}

Indoor experiments are conducted in the Toulouse Occitanie Drone Flight Arena at ENAC, Toulouse, France. The flight arena measures $10 \times 10$~meters, and all experiments are performed using a quadrotor equipped with the Paparazzi autopilot system \cite{Hattenberger2014}. The arena is equipped with a 16-camera OptiTrack motion capture system, which provides a highly accurate real-time state estimation of the vehicle.

During flight, the vehicle's position and velocity are  estimated by the OptiTrack system and transmitted via XBee radio modem and fused onboard with Paparazzi's IMU for full state estimation. Obstacle perception is performed exclusively onboard using the LiDAR sensor described in the previous section.

\begin{figure}
  \begin{subfigure}[b]{0.23\textwidth}
    \includegraphics[width=\linewidth]{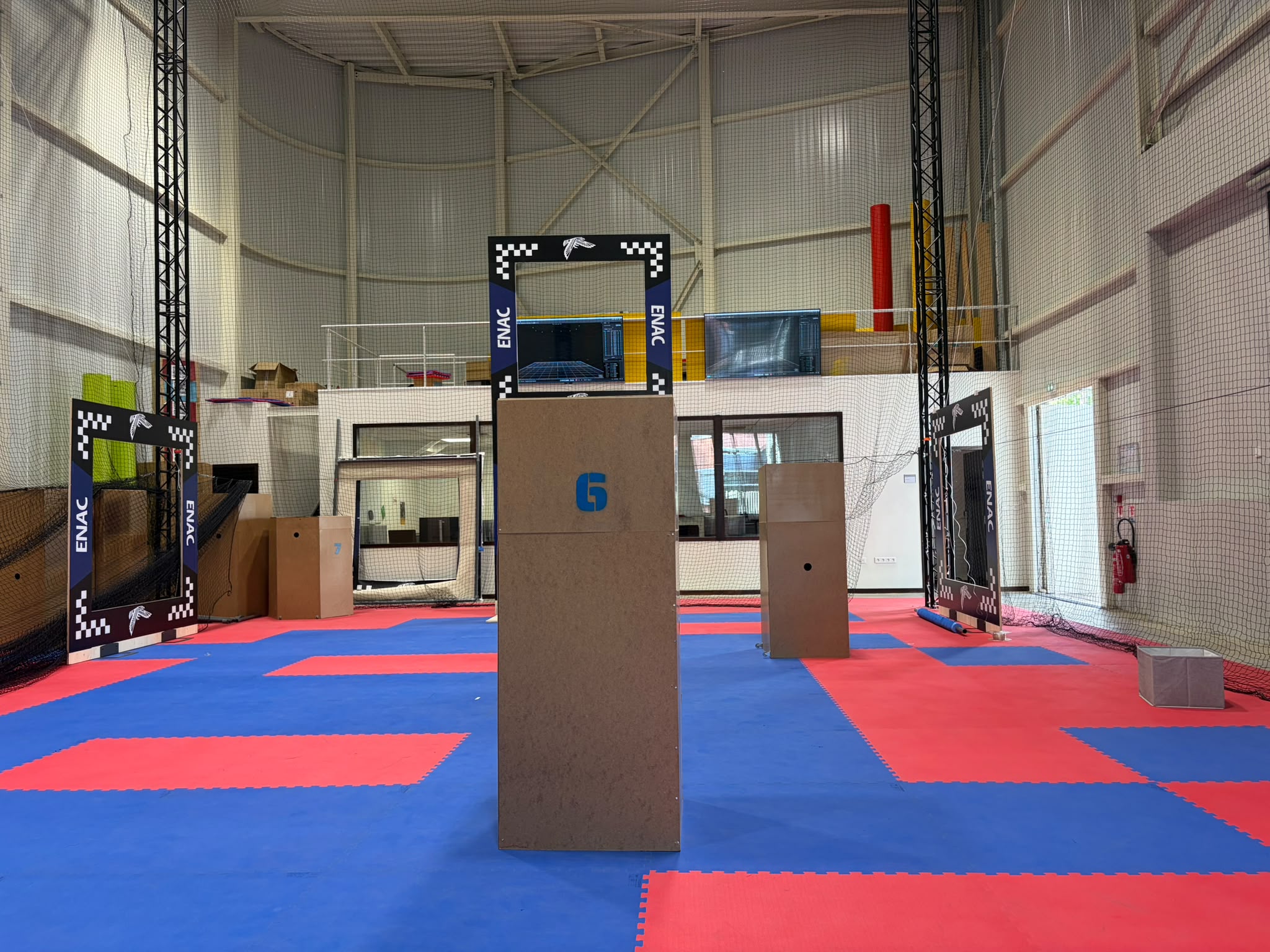}
    \caption{}
    \label{fig:arena1}
  \end{subfigure}
  \begin{subfigure}[b]{0.23\textwidth}
    \includegraphics[width=\linewidth]{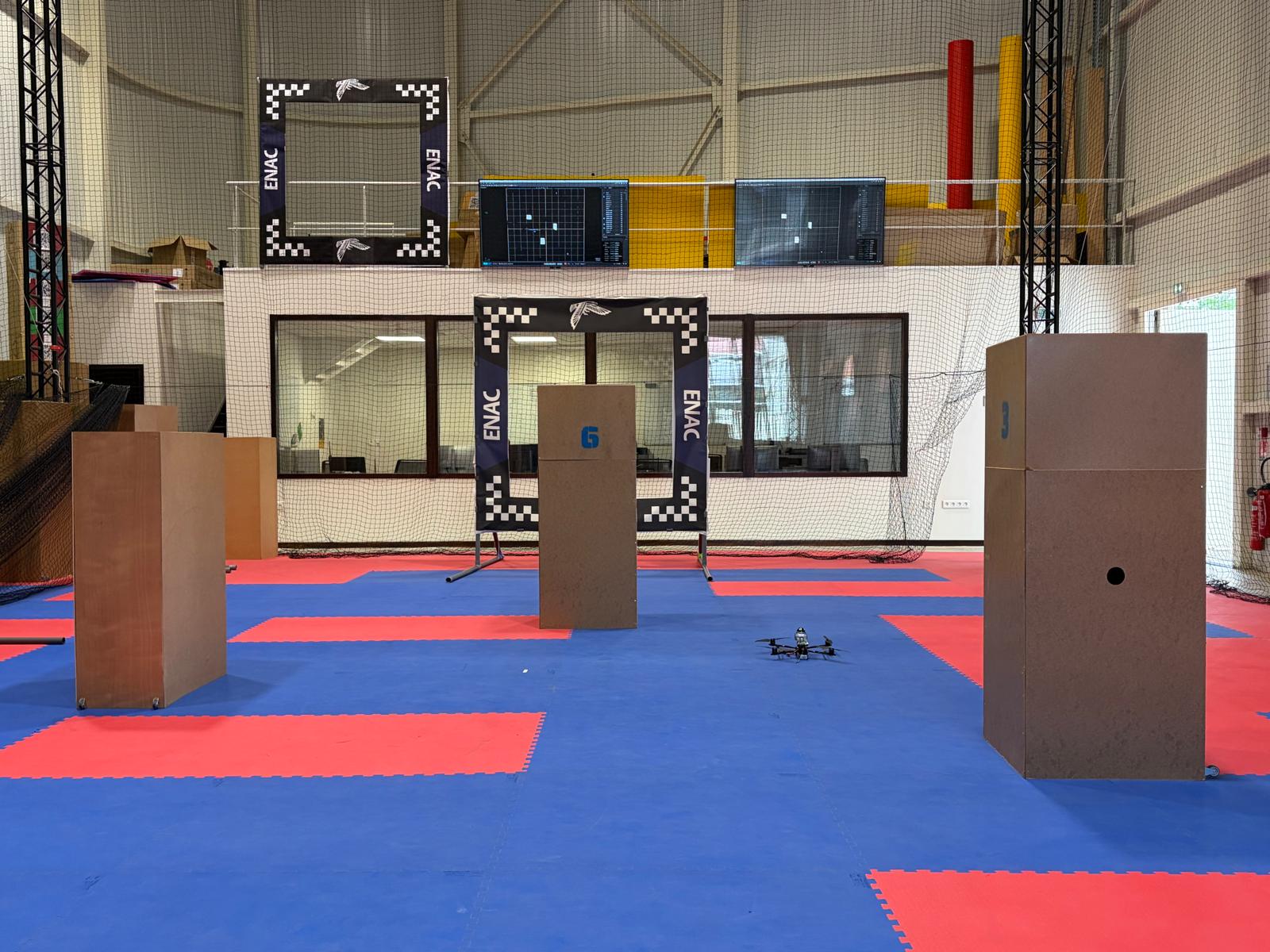}
    \caption{}
    \label{fig:arena2}
  \end{subfigure}
  \caption{Experimental setups for the go to waypoint, (a), and follow direction, (b) scenarios. On (a) two obstacles are placed between the waypoints the drone has to reach. On (b) the drone has to navigate around three obstacles while following the nominal direction command.}
  \label{fig:arenas}
\end{figure}

The obstacles are placed at fixed locations in the arena, as shown in Fig.~\ref{fig:arenas}. However, their positions are not known \textit{a priori} to the drone and are assumed to be non-static from the vehicle's perspective. Consequently, the drone continuously scans the environment and executes the obstacle detection algorithm at every time step.

The physical specifications of the vehicle can be found in Table \ref{tab:vehicle_specs}.

\begin{table}[h!]
  \centering
  \caption{Physical specifications of the vehicle used for the experiments.}
  \label{tab:vehicle_specs}
  \begin{tabular}{ll}
    \hline
    Specification        & Value \\
    \hline
    Mass                 & 1200~g \\
    Maximum thrust       & 40~N \\
    Battery              & 7000~mAh, 11.1~V (3-cell) \\
    Motors               & T-motor air gear 450 V2 \\
    Flight time          & $\sim$20~min \\
    LiDAR                & Livox Mid-360, 10~Hz \\
    Companion computer   & Raspberry Pi 5\\
    Autopilot            & Tawaki V2.0 (Paparazzi)\\
    \hline
  \end{tabular}
\end{table}

\begin{figure}
  \begin{subfigure}[b]{0.23\textwidth}
    \includegraphics[width=\linewidth]{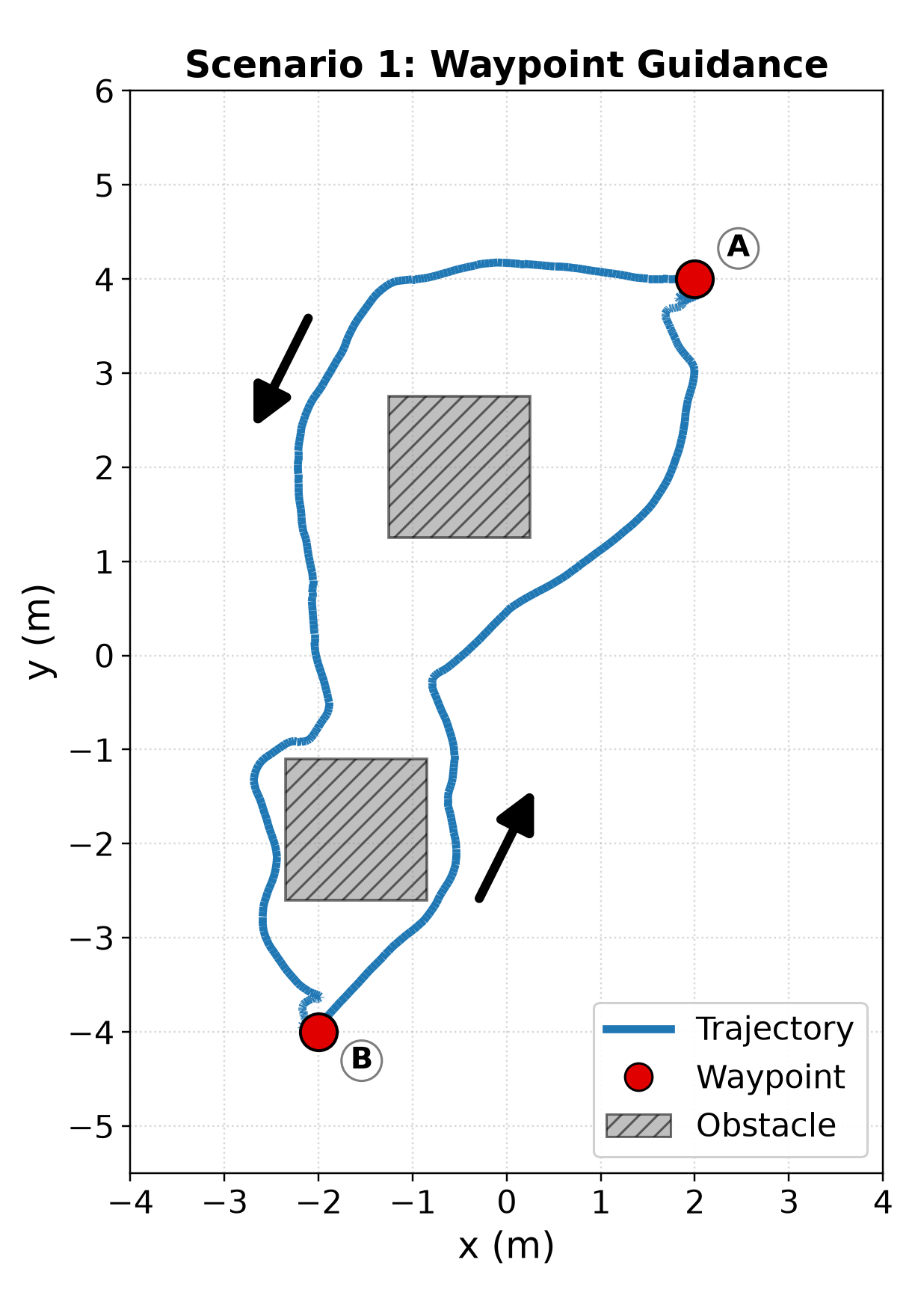}
    \caption{}
    \label{fig:waypoint}
  \end{subfigure}
  \begin{subfigure}[b]{0.23\textwidth}
    \includegraphics[width=\linewidth,page=2]{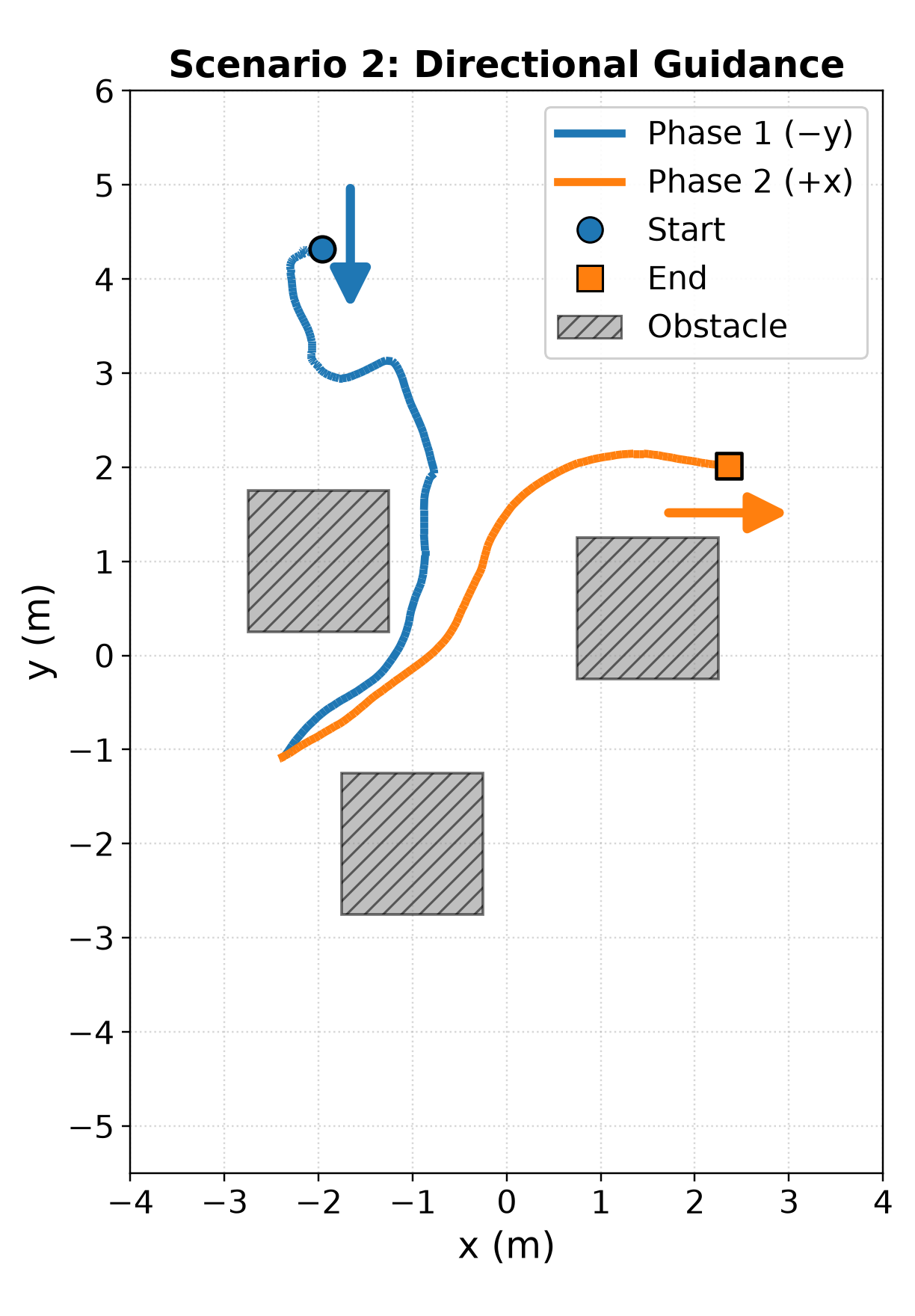}
    \caption{}
    \label{fig:direction}
  \end{subfigure}
  \caption{Experimental drone trajectories for the first, (a), and second, (b), scenarios. (a) The drone starts at waypoint A, flies to waypoint B, and returns to waypoint A while moving counter-clockwise (black arrows). The trajectory (blue) successfully avoids the obstacles (zebra-patterned squares). (b) For each coloured trajectory segment, the commanded direction is indicated by the arrow of the corresponding colour. The resulting trajectory successfully avoids the obstacles.}
  \label{fig:formations}
\end{figure}

\subsection{Scenario 1: Waypoint guidance}

In the first scenario, the vehicle is required to navigate from waypoint $A = (-2, -4)$~meters to waypoint $B = (2, 4)$~meters and return to the starting position. The chosen parameters for \eqref{eq:V} are $v_d = 0.5$ meters per second and $k_c = 1$. The maximum velocity is limited to $0.5$~meters per second for safety. Two obstacles are placed at $(-1.5, -1.75)$~meters and $(-0.5, 2)$~meters. 

The resulting trajectory is shown in Fig.~\ref{fig:waypoint}. The vehicle successfully reaches both waypoints while maintaining collision-free operation throughout the experiment.

\subsection{Scenario 2: Directional guidance}

In the second scenario, the vehicle is tasked with following a reference direction. The commanded direction is initially $(0,1)$ and is later changed to $(1,0)$. The chosen parameters for \eqref{eq:V} are $v_d = 0.5$ meters per second and $k_c = 1$. The maximum velocity is again limited to $0.5$~meters per second. Three obstacles are placed at $(-2,1)$~meters, $(1.5,0.5)$~meters, and $(-1,-2)$~meters. 

The resulting trajectory is shown in Fig.~\ref{fig:direction}. The vehicle is able to track the commanded direction while avoiding the obstacles.

\section{Discussion}

The experimental results demonstrate the feasibility of the proposed guidance system in unknown environments using only onboard LiDAR sensing. In both validation scenarios, the MAV successfully achieved its objectives, reaching the requested waypoints in the waypoint-navigation scenario and following the commanded direction in the directional-guidance scenario while avoiding all detected obstacles. However, the resulting trajectories exhibit noticeable oscillations and sharp turns rather than smooth paths. This behaviour is primarily attributed to instabilities in the obstacle detection algorithm. The estimated obstacles changed dimensions erratically between frames. These fluctuations caused corresponding variations in the repulsive forces generated by \textit{PGFlow}, leading to oscillatory trajectory corrections. 

The MAV operated at a relatively low speed of approximately $0.5,\mathrm{m/s}$. Further validation is required before evaluating the proposed guidance system at higher flight speeds. Additionally, all obstacles considered in the experiments were static. The current approach is not designed to handle fast-moving obstacles, as it relies solely on their instantaneous positions. To address this limitation, a future extension could incorporate an obstacle motion prediction module to estimate future obstacle positions.

Rather than continuously performing low-level obstacle avoidance, the operator is only required to specify a desired direction of travel while the vehicle autonomously reacts to the surrounding environment. Although no formal user study was conducted, this mode of operation has the potential to reduce operator workload and the level of piloting expertise required, making the approach attractive for applications such as search and rescue, where personnel may not have extensive experience operating MAVs.

The proposed guidance algorithm is computationally efficient, provided that the number of obstacle panels remains limited. At each obstacle update, an $n \times n$ matrix inversion is required, resulting in a computational complexity of $O(n^3)$, where $n$ denotes the total number of panels. Consequently, the computational cost increases rapidly as additional panels are introduced. In practice, however, the number of panels can be kept small by removing redundant or distant panels while preserving an adequate approximation of the obstacle geometry. The \textit{PGFlow} algorithm runs comfortably at $100$ Hertz. As the number of panels increases, the maximum frequency at which the obstacle matrix can be computed decreases. The relationship between this frequency and the maximum allowed number of detected panels is summarized in Table \ref{tab:hzvspanels}. Even at higher frequencies the maximum allowed number of detected panels remains sufficiently high for reliable real-time obstacle avoidance.

\begin{table}[h]
    \centering
    \caption{Relation between the frequency at which the obstacle detection algorithm is run and the maximum number of detected panels allowed for real-time safe obstacle avoidance.}
    \begin{tabular}{|c|c|c|c|c|}
        \hline 
        Frequency (Hz) &  1 & 5 & 10 & 20 \\
        \hline 
        Maximum number of panels & 2025 & 1100 & 725 & 506 \\
        \hline
    \end{tabular}
    \label{tab:hzvspanels}
\end{table}

Obstacle perception proved to be one of the most challenging components of the system. The dense LiDAR point cloud makes it difficult to segment individual obstacles and group the corresponding points, resulting in an unstable obstacle estimation algorithm, where object dimensions would change erratically between frames. Furthermore, the complete point cloud is currently transmitted from the C++ LiDAR driver to the Python guidance module via TCP before any processing is performed. A more efficient implementation would perform the filtering and obstacle extraction directly within the C++ driver, reducing both communication overhead and computational cost. Machine-learning-based methods would probably be a better fit to convert the point cloud into concrete obstacles.

Several simplifying assumptions were made in this work. The vehicle state was obtained from an external OptiTrack motion capture system, eliminating the need for onboard state estimation. Furthermore, guidance was only considered in the horizontal plane, with motion along the $z$-axis being neglected. Consequently, the experimental validation focuses solely on the proposed guidance strategy.

Only convex obstacles were considered. Concave obstacles are more challenging to navigate around and may introduce stagnation points (local minima) in the guidance field. In practice, concave obstacles can be approximated using multiple convex panels, albeit at the cost of approximation accuracy.

Finally, the proposed system was validated using a single MAV. Extending the approach to multi-vehicle operation is a natural continuation of this work.

\section{Conclusion}

This paper presented a fully onboard integrated guidance system for MAV navigation in unknown environments, addressing two scenarios: convergence to user-defined waypoints and tracking of a commanded direction. Obstacle detection is performed using an onboard LiDAR sensor, while obstacle avoidance is ensured by a guidance algorithm \cite{bilgin_urban_2024, fahimi_obstacle_2009}. The nominal motion is generated using a guiding velocity vector field approach \cite{Machado2026}.

The system was experimentally validated in an indoor environment with unknown convex obstacles. In both scenarios, the MAV successfully achieved its objectives without encountering local minima, while maintaining low computational cost. It has not been tested with concave obstacles.

These results demonstrate the feasibility of fully onboard guidance and obstacle avoidance in unknown environments using only onboard sensing. This enables intuitive high-level safe control with minimal operator expertise.

\flushend
\bibliographystyle{unsrt}
\bibliography{imav_bibliography}
\raggedend

\end{document}